\pdfoutput=1

\documentclass[11pt]{article}

\usepackage{acl}

\usepackage{times}
\usepackage{latexsym}
\usepackage{url}
\usepackage{graphicx}
\graphicspath{ {./images/} }

\usepackage{mathabx}

\usepackage{marvosym}

\usepackage[T1]{fontenc}

\usepackage[utf8]{inputenc}

\usepackage{microtype}

\usepackage{booktabs}
\usepackage{multirow}
\usepackage{siunitx}
\usepackage{xspace}

\usepackage{makecell}

\newcommand{\minisection}[1]{\noindent{\bf #1}\hspace{0.6em}}

\newcommand{\dataset}{\texttt{Choice-75}}

\title{\texttt{Choice-75}: A Dataset on Decision Branching in Script Learning}

\author{
    Zhaoyi Joey Hou$^1$\thanks{$^\ast$Work done while Zhaoyi Joey Hou was at University of Pennsylvania.}, Li Zhang$^2$, Chris Callison-Burch$^2$ \\
    $^1$ University of Pittsburgh, $^2$ University of Pennsylvania \\
    \texttt{joey.hou@pitt.edu}, \texttt{zharry@upenn.edu} 
}

\begin{document}
\maketitle

\begin{abstract}
Script learning studies how stereotypical events unfold,  enabling machines to reason about narratives with implicit information. Previous works mostly consider a script as a linear sequence of events while ignoring the potential branches that arise due to people's circumstantial choices. We hence propose \texttt{Choice-75}, the first benchmark that challenges intelligent systems to make decisions given descriptive scenarios, containing 75 scripts and more than 600 scenarios. We also present preliminary results with current large language models (LLM). Although they demonstrate overall decent performance, there is still notable headroom in hard scenarios.
\end{abstract}


\section{Introduction}
Events are the fundamental building blocks of the world around us. To understand the world, one has to comprehend the ways events interconnect with each other. For the same reason, the understanding of events and their relationship is critical for any intelligent system. Reasoning about the event-to-event relationships has long been a community effort from a wide range of perspectives, including studies in temporal relationship \cite{zhou-etal-2021-temporal, zhang-etal-2020-reasoning} and hierarchical relationship \cite{li-etal-2020-connecting, zhou-etal-2022-show}, both of which contribute to script generation \cite{chambers-jurafsky-2008-unsupervised, lyu-etal-2021-goal}. These tasks are challenging because event relations are often implicit and require commonsense to be uncovered. 

As an important direction of event-centric reasoning, script learning studies how stereotypical events unfold, which provides us with a human-centered perspective of events. The notion of scripts dates back to \citet{schank_scripts_1977}; since then, researchers have explored various aspects and applications of script learning, including narratives \cite{chambers-jurafsky-2010-database}, news events \cite{du-etal-2022-resin}, and instructions \cite{zhou-etal-2022-show}. These studies jointly demonstrate the promising nature of script learning in building better intelligent systems. 

\begin{figure}[t!]
    \centering
        \includegraphics[width=0.48\textwidth]{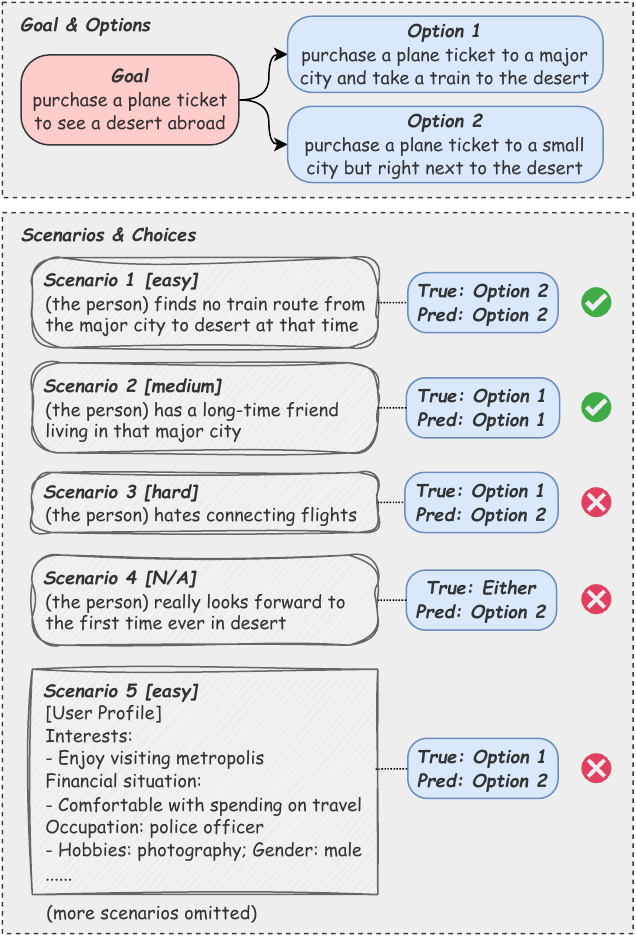}
    \caption{An example of \dataset{}. Each \texttt{goal-option} pair has multiple scenarios.}
    \label{fig:dataset}
    \vspace{-5mm}
\end{figure}

However, most of these previous works in script learning only consider scripts as linear developments of events.  In the real world, scripts include many crossroads where the next event can unfold in multiple ways. When a human acts as the agent, they would decide the direction to which a script branches. There has yet been no benchmark that challenges an intelligent system to model such a decision-making process. Therefore, we define and study such a decision branching task, as follows: given a particular scenario, an intelligent system needs to identify the more reasonable among two given options. One such example is in Figure \ref{fig:dataset}: given a scenario that \textit{the person finds no train route from the major city to desert at that time}, it would be obvious that the first option \textit{purchase a plane ticket to a major city and take a train to the desert} would not be feasible and the second \textit{purchase a plane ticket to a small city but right next to the desert} is the preferred answer.



We propose the first dataset, \dataset,  targeting such decision branching in scripts with 75 examples each with one goal and two options. Beyond that, we also collect more than 600 scenarios, with difficulty levels based on human judgment, and corresponding optimal choices. During dataset collection, we follow \citet{liu-etal-2022-wanli} and apply the human-in-the-loop paradigm to generate challenging examples.
We then experiment with state-of-the-art (SoTA) LLMs, including \texttt{text-davinci-003} and \texttt{gpt-3.5-turbo} and find that the level of performance of LLMs aligns with the difficulty levels based on human judgment. While these SoTA models demonstrate decent performance, there is still notable headroom in the hard cases. Our dataset would hopefully fuel further studies in AI-powered decision-making.\footnote{Dataset and code can be found at \url{https://github.com/JoeyHou/branching}.}

        


\begin{table}
\small
\centering
    \begin{tabular}{ccccc}
        \toprule
        \textbf{Format} & \textbf{Easy} & \textbf{Medium} & \textbf{Hard} & \textbf{Either}\\
         \midrule
        \makecell{Verb Phrase \\ (Manual)} & 65 & 76 & 36 & 65  \\
        \midrule
        \makecell{Verb Phrase \\ (Machine)} & 46 & 41 & 22 & 50 \\
        \midrule
        User Profile & 53 & 76 & 17 & 73  \\
         \midrule
        All & 164 & 193 & 75 & 188 \\
        \bottomrule
    \end{tabular}
    \caption{Counts of \texttt{scenario} in \dataset{}. 
    }
    \vspace{-4mm}
\label{table:level_dist}
\end{table}

\section{Dataset}
\label{sec:dataset}
\subsection{Overview}
\label{subsec:dataset_overview}
We begin by defining the basic unit of our dataset. Every data point in \dataset{} has the following: a \texttt{goal}, two options (\texttt{option-1} and \texttt{option-2}), a list of \texttt{scenario}, and a list of ground-truth \texttt{choice}, all of which in plain text. A \texttt{choice} could be \texttt{option-1}, \texttt{option-2}, or \texttt{either} (either option makes little difference under that \texttt{scenario}). For example, in scenario \texttt{\#4} in Figure \ref{fig:dataset}, both options would have little impact in achieving the \texttt{goal}, and thus the ground truth answer is \texttt{either}.


We use proScript \cite{Sakaguchi2021-ez} as the starting point for our dataset. It has 6.4k scripts that describe the sequence of actions for typical day-to-day activities, making it a suitable pool of goals for our task. We randomly sample 75 actions from proScript as the \texttt{goal} and manually write two feasible \texttt{option} to execute it. The \texttt{options} are written by one researcher and verified by two other researchers. In this way, we collect 75 \texttt{(goal, option-1, option-2)} tuples. 

After getting the feasible options for each \texttt{goal}, we add \texttt{scenario} and corresponding ground-truth \texttt{choice}. There are two data collection schemes for \texttt{scenarios}: manual writing by one researcher in this field (Section \ref{subsec:dataset_manual}) and human-in-the-loop scenario generation by an LLM (Section \ref{subsec:dataset_hitl}). To verify the quality of \texttt{scenarios} and corresponding \texttt{choices}, we randomly sample 290 \texttt{scenarios} and conduct an annotator agreement analysis on the ground-truth \texttt{choice}. The Fleiss' kappa coefficient for this sample is 0.59, which means moderate to substantial agreement \cite{rucker2012measuring}. More details about annotator agreements are in Appendix \ref{sec:appendix_iaa}.

After we finish collecting all the scenarios, we also define and annotate the difficulty level of each scenario in terms of how complex it is for a human to get the correct option choice. The criteria we use is the number of ``hops'' that the reasoning involves. In this way, we can explore multi-hop reasoning scenarios as a subset of our task. We defined four levels: \textit{easy, medium, hard}, and \textit{either} (for those scenarios without an optimal choice), with detailed discussions in Section \ref{subsec:diff_level}.

\subsection{Difficulty Level}
\label{subsec:diff_level}

Difficulty levels are based on the number of reasoning steps required for the correct option. Consider the \textit{library hours} example in Table \ref{table:level_diff}.
\begin{table}[t!]
    \centering
    \small
    \begin{tabular}{p{70mm}}
        \toprule
        \textbf{\textit{Goal}}: find out the library's hours \newline
        \textbf{\textit{Option 1}}: call the library \newline
        \textbf{\textit{Option 2}}: search online for the library's hours  \\
        \toprule
        \textbf{\textit{Easy Scenario:}} have no internet connection \newline
        \textbf{\textit{Choice}}: Option 1 \\
        \midrule
        \textbf{\textit{Medium Scenario:}} have special requests about the book \newline
        \textbf{\textit{Choice}}: Option 1  \\
        \midrule
        \textbf{\textit{Medium Scenario (User Profile):}} \newline 
        Name: Doe; Interests: American history  \newline 
        Special circumstances: has a bad sore throat \newline
        ... (more details omitted) \newline
        \textbf{\textit{Choice}}: Option 2  \\
        \midrule
        \textbf{\textit{Hard Scenario:}} is 3 am in the morning \newline
        \textbf{\textit{Choice}}: Option 2    \\
        \bottomrule
    \end{tabular}
    \caption{Different levels in the \textit{library hours} case}
    \label{table:level_diff}
    \vspace{-3mm}
\end{table}

\minisection{Easy} In this level, scenarios explicitly refer to one option, directly or indirectly. Only one easy reasoning step is required for such decision-making. For example, ``internet connection'' is directly related to ``search online'' and makes it infeasible. 

\minisection{Medium} In this level, scenarios implicitly refer to one option, directly or indirectly. The level of simplicity is low, i.e. it is easy to relate based on commonsense. For example, ``special requests'' implies that the person needs to talk to a staff member, which is related to ``call the library''; for the same reason, ``has a very
bad sore throat'' implies that the person cannot talk, which is related to ``call the library''.

\minisection{Hard} In this level, scenarios implicitly refer to something related to one option. These scenarios typically require the combination of commonsense knowledge and multiple steps of reasoning. For example, one needs to know that ``3 a.m. in the morning'' implies that the library is very likely to be closed; then one needs to further reason that in a closed library, no one would pick up the phone. This makes ``call the library'' infeasible. 

\subsection{Manual Scenario Annotation}
\label{subsec:dataset_manual}
The manual-written scenarios are all in verb phrase format, for example, scenario \texttt{\#1} to \texttt{\#4} in Figure \ref{fig:dataset}. In some cases, the scenario describes an event, e.g., ``finds no train route from the major city to desert at that time'' (scenario \texttt{\#1}); in other cases, the scenario describes a state of a person, either concrete or abstract, e.g., ``hates connecting flights'' (scenario \texttt{\#3}). Summary statistics about manual scenario generation are in Table \ref{table:level_dist}. 

\subsection{Human-in-the-Loop Generation}
\label{subsec:dataset_hitl}
During the manual scenario generation, coming up with high-quality hard scenarios requires a significant amount of mental effort. Therefore, we use a human-in-the-loop data generation paradigm and create two additional subsets of hard scenarios. The first subset is also in verb phrase format (same as the manual-written ones) and is referred to as \textit{machine-generated verb phrases}; the second subset comes in a different format, i.e. user profile in a bullet-point format, referred to as \textit{user profiles}. 

In terms of data collection procedure, we follow \cite{liu-etal-2022-wanli} by these steps\footnote{We skip the automatic filtering because the level of challenge is very hard to automatically measure.}: first, collect a series of challenging scenarios as exemplars; then, over-generate similar scenarios by few-shot prompting an LLM; lastly, manually review and curate the generated scenarios to ensure their validity. Note that, although the initial goal for this step is to create as many hard scenarios as possible, during the manual review and curation step, we still find many machine-generated scenarios that are not hard. Instead of assuming all the machine-generated scenarios are hard, we annotate their difficulty levels based on the same criteria, with the same annotator setup, described in Section \ref{subsec:diff_level}. 

\begin{figure}[t!]
    \centering
        \includegraphics[width=0.48\textwidth]{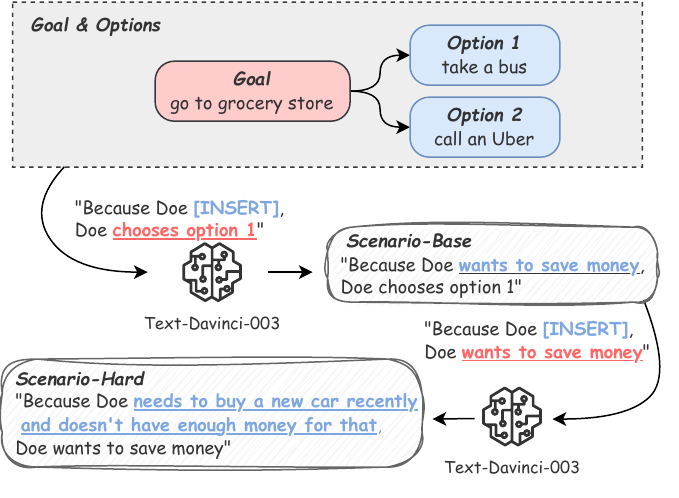}
    \caption{Hard scenario generation (verb phrase)}
    \label{fig:dataset_verb_phrase}
    \vspace{-1mm}
\end{figure}

\minisection{Verb Phrase}
The first type of hard scenario is the same as the manual written format, verb phrases. For the over-generation step, instead of a few-shot generation, we do a two-step prompting to simulate multi-hop reasoning (Figure \ref{fig:dataset_verb_phrase}). We first prompt a \texttt{text-davinci-003} model to generate a scenario that leads to one choice (i.e. \texttt{scenario-base}); then we do another few-shot prompting to generate a new scenario that leads to the \texttt{scenario-base} and save it as \texttt{scenario-hard}. The \texttt{scenario-hard} then goes through manual review and curation. More details are in Appendix \ref{sec:appendix_data_generation}.
\newline

\begin{figure}[t!]
    \centering
        \includegraphics[width=0.48\textwidth]{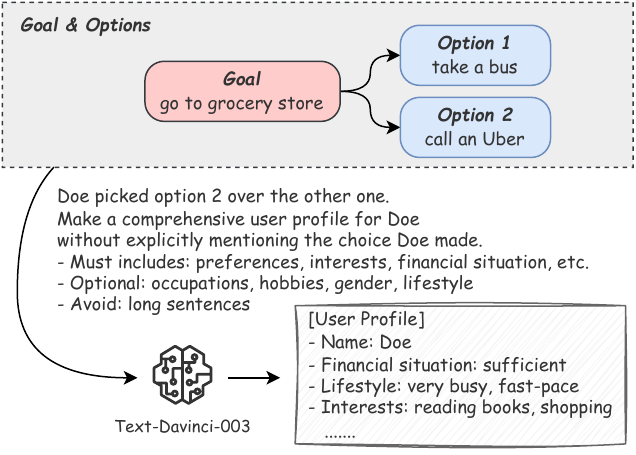}
    \caption{Hard scenario generation (user profile)}
    \label{fig:dataset_user_profile}
    \vspace{-4mm}
\end{figure}


\minisection{User Profile} 
Another type of hard scenario is a user profile in the form of an unordered list, for example, scenario \texttt{\#5} in Figure \ref{fig:dataset}. Our consideration of user profiles in addition to standard textual contexts is motivated empirically. First, many smart assistant software needs to be personalized to assist user decision-making. Moreover, user profiles are closer to real-life situations where the traits of a user are mined from heterogeneous data sources rather than from short texts. Such profiles inevitably include noise, making the task more challenging. For the example above, the only relevant information to predict the optimal choice (\textit{Option 2}) is that Doe \textit{enjoys visiting metropolis}. 

In the over-generation step of user profile scenarios, we prompt a \texttt{text-davinci-003} model to generate a user profile that prefers one choice over another (Figure \ref{fig:dataset_user_profile}). In the prompt, we specify some hints and requirements for the output. For example, we require the model to include preferences, and financial situations, and make occupations, hobbies, and gender optional. These generated user profiles also go through human review and curation. More details are in Appendix \ref{sec:appendix_data_generation}.

\begin{table*}
\small
\setlength{\tabcolsep}{3pt}
\centering
  \begin{tabular}{cc|cc|cc|cc|cc|cc|cc}
    \toprule
    \multirow{2}{*}{\textbf{Group}} &
    \multirow{2}{*}{\textbf{Prompt}} &
      \multicolumn{2}{c}{\textbf{All}} &
      \multicolumn{2}{c}{\textbf{Binary}} &
      \multicolumn{2}{c}{\textbf{Easy}} &
      \multicolumn{2}{c}{\textbf{Medium}} &
      \multicolumn{2}{c}{\textbf{Hard}} &
      \multicolumn{2}{c}{\textbf{Either}} 
      \\
      && {003} & {Turbo} & {003} & {Turbo} & {003} & {Turbo} & {003} & {Turbo} & {003} & {Turbo}& {003} & {Turbo} \\
      \midrule
        \multirow{2}{*}{\textbf{\shortstack{Verb Phrase \\ (Manual)}}} 
            & naive & 0.60 & 0.63 & 0.81 & 0.82 & 0.91 & 0.92 & 0.83 & 0.80 & 0.58 & 0.67 & 0.05 & 0.14 \\
            & story & 0.63 & \textbf{0.64} & \textbf{0.86} & 0.81 & \textbf{0.95} & 0.88 & \textbf{0.87} & 0.81 & \textbf{0.69} & \textbf{0.69} & 0.02 & \textbf{0.18} \\

        \midrule
        \multirow{2}{*}{\textbf{\shortstack{Verb Phrase \\ (Machine)}}} 
            & naive & \textbf{0.56} & \textbf{0.56} & 0.77 & \textbf{0.80} & 0.79 & 0.79 & 0.77 & \textbf{0.85} & 0.69 & \textbf{0.75} & \textbf{0.21} & 0.15 \\
            & story & 0.55 & 0.55 & 0.79 & \textbf{0.80} & 0.79 & \textbf{0.82} & \textbf{0.85} & 0.81 & 0.69 & \textbf{0.75} & 0.15 & 0.13 \\

        \midrule
        \multirow{2}{*}{\textbf{User Profile}} 
            & naive & \textbf{0.61} & 0.59 & 0.72 & 0.69 & \textbf{0.78} & 0.73 & 0.73 & 0.69 & 0.47 & \textbf{0.60} & \textbf{0.40 }& \textbf{0.40} \\
            & story & 0.50 & 0.60 & 0.57 & \textbf{0.73} & 0.58 & 0.76 & 0.60 & \textbf{0.74} & 0.40 & \textbf{0.60} & 0.37 & 0.34 \\

        \midrule
        {\textbf{Average}} 
            & & 0.57 & \textbf{0.60} & 0.75 & \textbf{0.77} & 0.80 & \textbf{0.82} & 0.77 & \textbf{0.78} & 0.59 & \textbf{0.68} & 0.20 & \textbf{0.22} \\
           
    \bottomrule
    \hline
  \end{tabular}
  
  \caption{Prediction accuracy by difficulty levels. \textbf{Binary}: overall performance on binary classification (i.e. \texttt{Option 1} or \texttt{Option 2}); \textbf{All}: overall performance on three-class classification.}
  \label{table:result}
  \vspace{-3mm}
\end{table*}

\section{Method and Experiments}
\label{sec:experiments}

Out of the 75 data points in \dataset{}, we randomly hold out 10 data points as demonstrations for in-context learning and the rest for evaluation.

We formulate the task of predicting optimal choice as an in-context learning task: the \texttt{goal}, two \texttt{option}, and one \texttt{scenario} are presented in the prompt; an LLM is then responsible for completing the prompt with the optimal \texttt{choice} (or \texttt{either}). The few-shot context consists of 9 demonstrations with the same format, including 3 different choices and 3 difficulty levels. 

We include two models: \texttt{text-davinci-003} and \texttt{gpt-3.5-turbo} \footnote{Our last experiment was in 05/2023; the closest variant of turbo model is \texttt{gpt-3.5-turbo-0613}}. We set temperature to 0, max\_tokens to 30, top\_p to 1, presence\_penalty to 0, and ferquency\_penalty to 0. We also provide two prompt formats: naive prompt and story prompt. Prompt templates are in Appendix \ref{sec:appendix_b}.


\section{Results and Analysis}
\label{sec:analysis}

\subsection{Difficulty Levels}
The most outstanding result is the alignment of human judgment of difficulty and the model's performance. As shown in Table \ref{table:result}, there is an obvious gap between easy, medium, and hard scenarios across every setting. Although the models we test demonstrate decent performance in easy and medium levels, hard and either scenarios remain challenging. This again demonstrates that LLMs struggle more in multi-hop reasoning.

\subsection{Human Performance}
We also conduct a human performance analysis on a subset of the dataset with 290 sampled scenarios, each answered by two participants. The average human accuracy is 0.74, compared to 0.60 from the best model performance; the human accuracy is 0.76 for ``hard'' scenarios and 0.53 for ``either'' scenarios, both notably higher than the best model performances (i.e. 0.68 for  ``hard'' and 0.22 for ``either''). More details are in Appendix \ref{sec:appendix_human_performances}.

\subsection{Case Studies}
We take out one example from \dataset{} (see Figure \ref{fig:dataset}) and examine the performance of one model setup (\texttt{gpt-3.5-turbo} with \textit{story prompt}). For scenario \texttt{\#3}, the model fails to recognize that a small city usually requires a flight connection. For scenario \texttt{\#5}, a user profile example, although the scenario explicitly describes this person as \textit{``enjoy visiting metropolis''}, the model still gets it wrong. We can observe similar errors in other data points, confirming the challenge of the long context window and unrelated information introduced by the user profile format. More qualitative analyses are in Appendix \ref{sec:appendix_error_analysis}.



\section{Related Work}
\label{sec:related_work}
\minisection{Event-centric reasoning} and script learning \cite{schank_scripts_1977} are crucial domains of machine reasoning. Past efforts include procedure reasoning \cite{dalvi-etal-2019-everything, zhang-etal-2020-reasoning, zhou-etal-2022-show}, entity tracking \cite{tandon-etal-2020-dataset, zhang-etal-2023-causal}, and script learning \cite{chambers-jurafsky-2008-unsupervised, lyu-etal-2021-goal, Sakaguchi2021-ez}. All of these works above focus on singular chains of events while we focus on branching structures in events. 

In addition, a series of other works have explored the effect of a scenario or additional context on a given, main event. For example, \cite{rudinger-etal-2020-thinking} explores the influence of different scenarios on human interpretation of events, \cite{otani-etal-2023-textual} focuses on conversational tasks and analyzes the influence of different scenarios on human behaviors, and \cite{wang-etal-2023-cola} studies the context-dependency of event causality.

\minisection{Human decision-making} has been studied under single-agent and multi-agent settings. Efforts in the former focus on specific domains, such as financial earnings call \cite{keith-stent-2019-modeling}, online review text \cite{wang-etal-2019-human}, and fantasy text-adventure game \cite{qiu-etal-2022-towards}. In contrast, our methods and findings are more general. Efforts in the latter focus on dialogues and conversational AIs, such as dialogues \cite{bak-oh-2018-conversational, karadzhov-etal-2022-makes, fernandez-etal-2008-modelling} with an emphasis on modeling the differences among characters, which is not our focus.

\minisection{Human-in-the-loop dataset creation}
has been used for efficient data collection and quality improvement. Recent work shows that LLMs can effectively generate data for NLP tasks, including natural language inference \cite{liu-etal-2022-wanli}, structural data synthesis \cite{yuan2022synthbio}, script construction \cite{zhang-etal-2023-human}, hate speech detection \cite{tekiroglu-etal-2020-generating}. In our work, we closely follow the paradigm of \cite{liu-etal-2022-wanli} in dataset creation.


\section{Conclusion}
We investigate the decision-making ability of current SoTA LLMs and find room for improvement in hard decision-making scenarios when compared with human performance. We also observe a notable alignment between human judgment of difficulty and corresponding LLM performance. With the \dataset{} dataset, we introduce a new machine reasoning task where a model needs to incorporate implicit commonsense knowledge into decision-making. We hope this task can be a starting point for future studies of LLM's capability of daily decision-making.

\section*{Limitations}
The first and most obvious drawback of \dataset{} is its distribution. Since we build \dataset{} from the \textit{steps} from \texttt{proScript} \cite{Sakaguchi2021-ez}, which focuses on daily procedures; therefore the distributions of word choices, writing styles, and domains are inherently limited. Therefore, specific adaptation would be required if the data come from a different domain. 

Secondly, the size of the dataset is also relatively small due to limited annotation resources available to us. This also brings potential biases from the annotator, although we try to address this issue by having another annotator verify the annotations. Such a bias in the dataset might negatively impact the models fine-tuned on our dataset in the future. That could potentially lead to inappropriate prediction results from those fine-tuned models if the end users are from a different cultural background.

In addition, in the \dataset{}, we make a lot of assumptions that are essentially oversimplified representations of real-world scenarios. For example, we assume each goal has two mutually exclusive choices, while in some cases there are much more choices (not \textit{two}) and each choice overlaps with others (not \textit{mutually exclusive}). There are lots of ways to expand and enrich this dataset and we leave this as future work.

Last but not least, we also do not conduct any prompt engineering due to a limited computation budget. We only experiment with two very basic prompt formats, a fixed number of few-shot samples, and a fixed set of GPT generation parameters. It would also be interesting for future works to study the performance of different language models and different prompt settings on \dataset{}.

\section*{Acknowledgements}
We thank Nathanael Chambers for inspiring this work and for valuable discussions. We also thank the help from Xiang Lorraine Li for her suggestions on revising this paper. The data annotation would not be possible without the help from Bhiman Kumar Baghel, Alejandro Ciuba, Qi Cheng, Weihang Gao, Xiang Lorraine Li, Yifei Couson Ning, Tai Nguyen, Zilin Shelly Ren, and Hainiu Xu (in alphabetical order).

This work is supported in part by the DARPA KAIROS Program (contract FA8750-19-2-1004), AFRL (contract FA8750-23-C-0507), the Office of the Director of National Intelligence (ODNI) via the IARPA HIATUS Program (contract 2022-22072200005), the NSF (Award 1928631), and gifts from Roblox and Salesforce. Approved for Public Release, Distribution Unlimited. The views and conclusions contained herein are those of the authors and should not be interpreted as necessarily representing the official policies, either expressed or implied, of DARPA, ODNI, IARPA, NSF, AFRL, the U.S. Government, or of Roblox or Salesforce. The U.S. Government is authorized to reproduce and distribute reprints for governmental purposes notwithstanding any copyright annotation therein.

\bibliography{custom}
\bibliographystyle{acl_natbib}

\appendix

\section{Inter-Annotator Agreement}
\label{sec:appendix_iaa}
We collected annotations for 290 randomly sampled scenarios from 7 researchers in total. For each scenario, the optimal choice (i.e. \texttt{Option 1}, \texttt{Option 2}, or \texttt{Either}) is annotated by 3 researchers. The overall Fleiss' kappa is 0.59, which lies on the borderline between moderate and substantial agreement. In particular, there are 125 verb phrases (manual) with Fleiss' kappa being 0.66; 65 verb phrases (machine) with Fleiss' kappa being 0.49; and 100 user profiles with Fleiss' kappa being 0.55. 

\section{Human-in-the-loop Data Generation Prompting Details}
\label{sec:appendix_data_generation}

There are three implementation details about the prompting setup for Human-in-the-loop data generation.\\
First, in all prompts, we include ``overall goal'', which is the goal for the script from \texttt{proScript}, while ``step goal'' is the goal the person needs to make a decision on as well as the goal we refer to in the paper. We include the ``overall goal'' just to provide additional context information.\\
Second, for all prompts, the results would be the scenarios with the correct answer being \textit{option 1}. We also swap two options in these prompts so that we can get hard scenarios with the correct answer being \textit{option 2}.\\
Third, for all prompts, we provide four hand-written demonstrations, all of which come from the 10 held-out training scripts described in Section \ref{sec:experiments}. We use the insertion mode of the provided OpenAI API, \texttt{text-davinci-003} as the model, and 0.75 as the temperature.\\

\minisection{Verb Phrase}
\begin{quote}
    \textbf{\textit{Prompt Step 1:}}\\
    Doe wants to go \{overall goal\}. One of the steps towards that is \{step goal\}. Doe has two options: 1) \{option 1\} or 2) \{option 2\}\\
    Because Doe [INSERT], Doe chooses option 1.\\
    \textbf{\textit{Prompt Step 2:}}\\
    Doe wants to \{overall goal\}. One of the steps towards that is to \{step goal\}. Doe has two options: 1) \{option 1\} or 2) \{option 2\}\\
    Because [INSERT], Doe \{scenario-base\}. Therefore, option 2 is not available or not desirable for Doe and Doe chooses option 1.
\end{quote}
\minisection{User Profile}
\begin{quote}
\textbf{\textit{Prompt:}}\\
A person Doe would like to \{overall goal\} and need to finish the step of \{step goal\}. Doe now has two options: option 1 is to \{option 1\} and option 2 is to \{option 2\}. Eventually, Doe picked option 1 over the other. \\
Make a comprehensive user profile for Doe without explicitly mentioning the choice Doe made.\\
Must-includes: preferences, interests, financial situation, etc.\\
Optional: occupations, hobbies, gender, lifestyle\\
Avoid: long sentences\\
User Profile:
\end{quote}



\section{Decision Prediction Prompting Details}
\label{sec:appendix_b}

During inference time, we provide 9 in-context demonstrations, which are the combination of 3 difficulty levels and 3 labels. We also set the temperature to 0 to ensure consistency across runs. \\ \\
\minisection{Naive Prompt}
\begin{quote}
{[}Goal{]}: \{step goal\}\\
{[}Option 1{]}: \{option 1\}\\
{[}Option 2{]}: \{option 2\}\\
{[}Scenario{]}: \{scenario\}\\
{[}Question{]}: Given the Scenario, which option above is the better choice in order to achieve the Goal?\\
1) Option 1\\
2) Option 2\\
3) Either one, since they have similar effect when it comes to the goal\\
{[}Answer{]}:
\end{quote}

\minisection{Story Prompt}
\begin{quote}
A person Doe needs to \{step goal\}. Now there are two options for Doe: we can either \{option 1\} (Option 1) or \{option 2\} (Option 2). \\
Suppose Doe \{scenario\}.\\
{[}Question{]}: Given the Scenario, which option above is the better choice in order to achieve the Goal?\\
1) Option 1\\
2) Option 2\\
3) Either one, since they have similar effect when it comes to the goal\\
{[}Answer{]}:
\end{quote}

\begin{table}
\small
\centering
    \begin{tabular}{ccccc}
        \toprule
        \textbf{Format} & \textbf{Easy} & \textbf{Medium} & \textbf{Hard} & \textbf{Either}\\
         \midrule
        \makecell{Verb Phrase \\ (Manual)} & 0.94 & 0.81 & 0.82 & 0.62  \\
        \midrule
        \makecell{Verb Phrase \\ (Machine)} & 0.94 & 0.77 & 0.68 & 0.41 \\
        \midrule
        User Profile & 0.89 & 0.78 & 0.75 & 0.53  \\
         \midrule
        All & 0.92 & 0.79 & 0.76 & 0.53 \\
        \bottomrule
    \end{tabular}
    \caption{Human performance (accuracy) on \dataset{}}
    \vspace{-6mm}
\label{table:human_performance}
\end{table}
\section{Human Performance}
\label{sec:appendix_human_performances}
We tested human performance on a subset of 290 samples (Table \ref{table:human_performance}). For some entries in easy and medium difficulty levels, there is not much difference between human and model performance. However, in the hard and ``either'' difficulty levels, there is notable headroom ahead of the language models tested in our experiments.

\section{Qualitative Error Analysis}
\label{sec:appendix_error_analysis}
Here we provided two qualitative analyses where the prediction is different from the ground truth answer:\\
\minisection{Example 1}
\begin{quote}
    - \textbf{\textit{Goal}}: purchase a plane ticket\\
    - \textbf{\textit{Option 1}}: purchase a plane ticket to a major city but far from the desert\\
    - \textbf{\textit{Option 2}}: purchase a plane ticket to a small city but right next to the desert\\
    - \textbf{\textit{Scenario}}: hate connecting flights\\
    - \textbf{\textit{Level}}: hard\\
    - \textbf{\textit{True Answer}}: option 1\\
    - \textbf{\textit{Predicted Answer}}: option 2
\end{quote}
\textbf{Analysis}: for the example above, a flight to a major city \& far from the desert would most likely require a connecting flight as the next step; a flight to a small city near the desert would be ideal since it does not require a connecting flight. The model is not able to conduct these reasoning steps given the output. \\
\newline \minisection{Example 2}
\begin{quote}
- \textbf{\textit{Goal}}: pack hiking backpacks \\
- \textbf{\textit{Option 1}}: bring process food for every meal \\
- \textbf{\textit{Option 2}}: bring raw foods and some cookware to cook at the campsite \\
- \textbf{\textit{Scenario}}: want to enjoy every minute of the holiday \\
- \textbf{\textit{Level}}: medium \\
- \textbf{\textit{True Answer}}: option 1 \\
- \textbf{\textit{Predicted Answer}}: option 2 
\end{quote}
\textbf{Analysis}: if the person brings raw foods and cooks them at the campsite, most likely they would have to spend more time on the cooking instead of enjoying the hike. Therefore option 1 is preferable.

\end{document}